\documentclass[sigconf]{acmart}

\usepackage{enumitem}
\usepackage{multirow}
\usepackage{tcolorbox}
\usepackage{listings}
\usepackage{algorithm}  
\usepackage{algpseudocode}  
\usepackage{amsmath}  
\usepackage{ragged2e}

\usepackage{subcaption}

\tcbuselibrary{breakable, skins}

\usepackage{stfloats}

\usepackage{balance}

\AtBeginDocument{%
  }

\copyrightyear{2026}
\acmYear{2026}
\setcopyright{cc}
\setcctype{by}
\acmConference[CIKM '26] {Proceedings of the 35th ACM International Conference on Information and Knowledge Management}{November 7--11, 2026}{Rome, Italy}
\acmBooktitle{Proceedings of the 35th ACM International Conference on Information and Knowledge Management (CIKM '26), November 7--11, 2026, Rome, Italy}
\acmISBN{979-8-4007-2539-5/2026/11}
\acmDOI{10.1145/3799682.3841120}
\begin{document}

\title{Incorporating Cognitive Load and Knowledge Transfer for Multi-Domain Knowledge Tracing}


\author{Haotian Zhang}
\affiliation{%
  \institution{State Key Laboratory of Cognitive Intelligence, University of Science and Technology of China}
  \city{Hefei}
  \country{China}}
\email{sosweetzhang@mail.ustc.edu.cn}

\author{Shucun Wang}
\affiliation{%
  \institution{State Key Laboratory of Cognitive Intelligence, University of Science and Technology of China}
  \city{Hefei}
  \country{China}}
\email{shucunwang@mail.ustc.edu.cn}

\author{Jinze Wu}
\affiliation{%
  \institution{iFLYTEK AI Research}
  \city{Hefei}
  \country{China}}
\email{hxwjz@mail.ustc.edu.cn}

\author{Liang Ding}
\affiliation{%
  \institution{iFLYTEK AI Research}
  \city{Hefei}
  \country{China}}
\email{liangding3@iflytek.com}

\author{Shuochen Liu}
\affiliation{%
  \institution{State Key Laboratory of Cognitive Intelligence, University of Science and Technology of China}
  \city{Hefei}
  \country{China}}
\email{shuochenliu@mail.ustc.edu.cn}

\author{Zhenya Huang}
\affiliation{%
  \institution{State Key Laboratory of Cognitive Intelligence, University of Science and Technology of China \& 
Institute of Artificial Intelligence, Hefei Comprehensive National Science Center}
  \city{Hefei}
  \country{China}}
\email{huangzhy@ustc.edu.cn}

\author{Jing Sha}
\affiliation{%
  \institution{iFLYTEK AI Research}
  \city{Hefei}
  \country{China}}
\email{jingsha@iflytek.com}

\author{Shijin Wang}
\affiliation{%
  \institution{State Key Laboratory of Cognitive Intelligence \& iFLYTEK AI Research}
  \city{Hefei}
  \country{China}}
\email{sjwang3@iflytek.com}
\correspondingauthor

\author{Qi Liu}
\affiliation{%
  \institution{State Key Laboratory of Cognitive Intelligence, University of Science and Technology of China \& 
Institute of Artificial Intelligence, Hefei Comprehensive National Science Center}
  \city{Hefei}
  \country{China}}
\email{qiliuql@ustc.edu.cn}

\renewcommand{\shortauthors}{Haotian Zhang et al.}

\begin{abstract}
Knowledge Tracing (KT) aims to assess students' dynamic knowledge states from their learning histories. While most existing KT methods focus on single-domain learning with notable success, real-world learning scenarios often involve multiple domains simultaneously, introducing two critical factors:  
1) Cognitive load, arising from managing learning across domains in both temporal and knowledge dimensions.  
2) Knowledge transfer, where knowledge states in one domain influence related states both within and across domains.  
In this paper, we focus on exploring these factors to improve students’ knowledge state assessment in multi-domain learning scenarios and propose a novel method incorporating cognitive \textbf{L}oad and knowledge \textbf{T}ransfer for \textbf{M}ulti-domain \textbf{K}nowledge \textbf{T}racing (\textbf{LT-MKT}).  
Specifically, to bridge isolated domains, LT-MKT first integrates textual information from questions and their associated concepts to construct a Multi-domain Hierarchical Graph, leveraging the advanced representational capabilities of large language models (LLMs).  
Then, cross-domain features in both the temporal and knowledge dimensions are explicitly modeled to capture the effects of cognitive load.  
Additionally, a knowledge transfer module is designed to model the propagation of knowledge states within and across domains.  
By jointly modeling these factors, LT-MKT enables more accurate prediction of students’ future performance.  
Finally, extensive experiments on real-world datasets demonstrate that our method achieves state-of-the-art performance.
Code is available at https://github.com/sosweetzhang/LT-MKT.
\end{abstract}

\ccsdesc[500]{Applied computing~E-learning}
\ccsdesc[500]{Social and professional topics~Student assessment}

\keywords{Knowledge Tracing, Cognitive Load, Knowledge Transfer}

\maketitle

\section{Introduction}

\begin{figure*}[t!]
  \centering
  \includegraphics[width=0.95\linewidth]{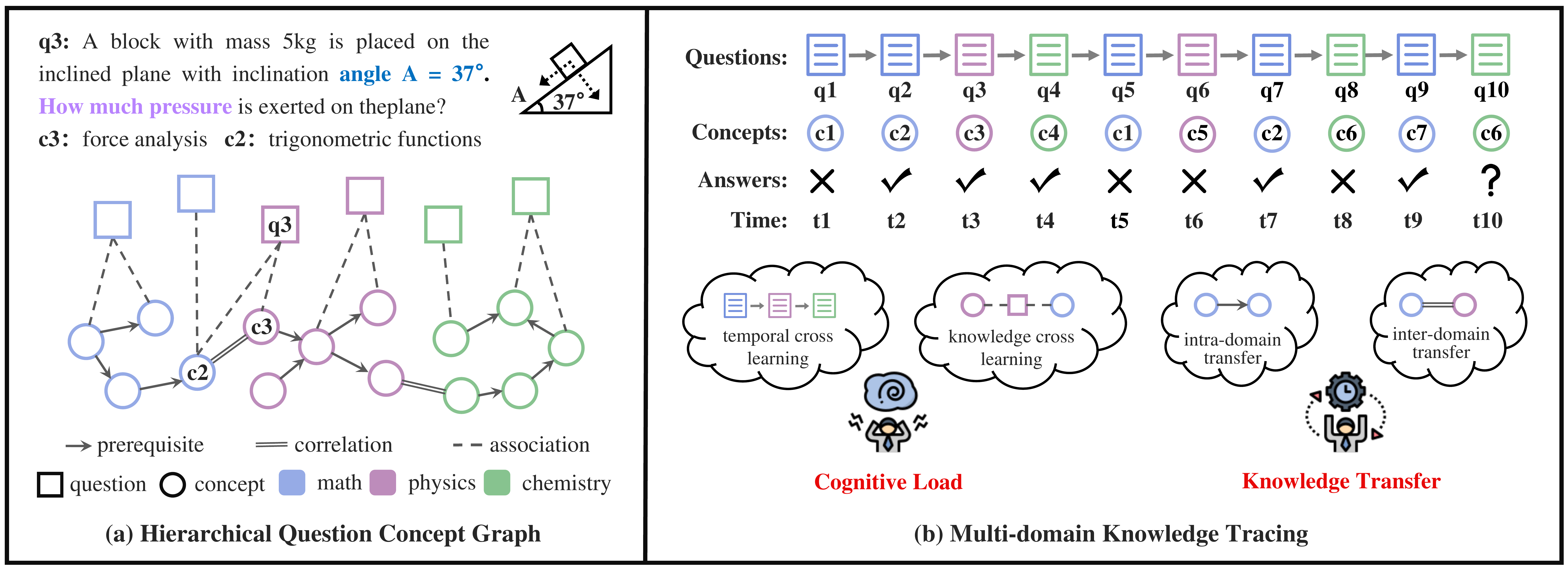}
  \vspace{-0.3cm}
  \caption{Illustration of Multi-domain Knowledge Tracing. (a) Hierarchical Question Concept Graph outlines relationships between concepts, as well as between questions and concepts. Questions in one domain (e.g., $q3$) may link to concepts from another (e.g., $c2$), and concepts across domains can also correlate (e.g., $c2$ and $c3$). (b) In multi-domain learning scenarios, students may alternate between domains like math, physics, and chemistry, enhancing learning through knowledge transfer but also increasing cognitive load.}
  \label{fig:intro}
   \vspace{-0.35cm}
\end{figure*}

Online learning has expanded rapidly, offering unmatched flexibility for both students and educators, enabling learning to occur anywhere, anytime~\cite{ktsurvey}. Platforms such as Coursera~\cite{Ku2011ITS} and ASSISTments~\cite{And2014Course} have demonstrated effectiveness by delivering intelligent, adaptive educational services~\cite{SONG2022KT}. 
These systems collect extensive data on student interactions, including exercise responses, enabling analysis of knowledge levels, learning preferences, and other key attributes. A central component of this analysis is Knowledge Tracing (KT), which aims to model and track students’ evolving knowledge states through their interactions with the system~\cite{ktsurvey,zhang2024item,zhang2026cbegrec}.

In the literature, extensive work has advanced knowledge tracing~\cite{shen2024KT}. Early methods relied on hidden markov models~\cite{BKT, DBKT} and logistic functions with various factors~\cite{PFA, LFA, KTM} to estimate students' knowledge states. The introduction of deep learning marked a significant shift, with Deep Knowledge Tracing (DKT) pioneering the application of neural networks to KT~\cite{DKT}. Since then, numerous variants have emerged, including memory-enhanced models~\cite{DKVMN,abdelrahman2019knowledge}, graph-based approaches~\cite{gkt,ni2023hhskt}, and attention-driven models~\cite{ghosh2020AKT,SAINT+}, etc. 
More recently, with the rapid development of large language models (LLMs), LLM-based approaches have been explored, leveraging their strong representation and reasoning capabilities to enhance knowledge state modeling, improve interpretability, and support knowledge-aware reasoning tasks~\cite{fu2024sinkt, han2025contrastive}.

Despite their success, most existing methods primarily focus on single-domain learning (e.g., math). However, many real-world learning scenarios, such as standardized exams like the GRE, require students to engage with multiple domains simultaneously. 
As illustrated in Figure~\ref{fig:intro}, multi-domain learning scenarios introduce additional complexity beyond single-domain settings, as a single question may simultaneously require knowledge from multiple domains. 
Such scenarios give rise to two critical factors that significantly affect students' knowledge states:
1) \textbf{Cognitive load}, resulting from learning across domains in both temporal and knowledge dimensions. 
Cognitive load theory~\cite{plass2010cognitive} posits that cognitive load originates from the \textit{complexity of learning materials and interactions}, which implicitly influence students' knowledge acquisition and the evolution of their knowledge states during the learning process.
In multi-domain learning scenarios, cognitive load mainly manifests in two forms:  
   \textit{Temporal dimension}: Students frequently switch between different domains over time. For instance, a student may study physics at time $t_3$, switch to mathematics at $t_4$, and then move to chemistry at $t_5$, leading to additional cognitive switching costs.  
   \textit{Knowledge dimension}: A single question may simultaneously require knowledge from multiple domains. For instance, solving the question $q_3$ at $t_3$ may require both $c_3$ force analysis from the physics domain and $c_2$ trigonometric functions from the mathematics domain, increasing the complexity of knowledge processing.
2) \textbf{Knowledge transfer}, where knowledge states in one domain influence related knowledge states both within and across domains. 
Transfer of learning theory~\cite{cormier2014transfer} emphasizes that knowledge concepts are inherently interconnected, and knowledge transfer occurs when \textit{understanding one concept influences the learning or understanding of related concepts}.
In multi-domain learning scenarios, knowledge transfer mainly occurs in two ways:  
   \textit{Intra-domain transfer}: Knowledge states propagate among related concepts within the same domain through prerequisite relationships, such as from addition to multiplication.  
   \textit{Inter-domain transfer}: Knowledge states transfer across different but related domains through semantic or functional correlations, such as between force analysis and trigonometric functions.
Therefore, accurately modeling knowledge states in multi-domain learning scenarios requires explicitly considering both cognitive load and knowledge transfer, which are typically overlooked in conventional single-domain KT methods.

In this paper, we focus on exploring the above factors to improve students’ knowledge state assessment in multi-domain learning scenarios, and propose a novel method incorporating cognitive \textbf{L}oad and knowledge \textbf{T}ransfer for \textbf{M}ulti-domain \textbf{K}nowledge \textbf{T}racing (\textbf{LT-MKT}). Unlike conventional single-domain KT methods that model knowledge states independently within each domain, LT-MKT explicitly captures cross-domain interactions from both cognitive and knowledge propagation perspectives.  
Specifically, to bridge isolated domains and establish semantic relationships across heterogeneous concepts, LT-MKT first integrates textual information from questions and their associated concepts to construct a Multi-domain Hierarchical Graph, leveraging the strong representational capabilities of large language models (LLMs). Based on this graph structure, LT-MKT further models cross-domain dependencies in both temporal and knowledge dimensions to characterize students' cognitive load during learning processes. In the temporal dimension, the model captures students' domain-switching behaviors over sequential interactions, while in the knowledge dimension, it models the cognitive burden introduced by questions requiring concepts from multiple domains simultaneously.  
Furthermore, to model the influence of related concepts on knowledge evolution, LT-MKT incorporates a knowledge transfer module to capture the propagation of knowledge states both within and across domains. The intra-domain transfer mechanism models prerequisite relationships among concepts within the same domain, whereas the inter-domain transfer mechanism captures semantic and functional correlations between concepts from different domains. By jointly modeling cognitive load and knowledge transfer, LT-MKT provides a more comprehensive representation of students’ learning processes and enables more accurate prediction of future performance.  
Finally, extensive experiments on real-world datasets demonstrate the effectiveness of LT-MKT, consistently achieving state-of-the-art performance compared with existing knowledge tracing methods.

Our main contributions are summarized as follows:
\begin{itemize}[leftmargin=*,itemsep=0.8pt]
    \item We highlight the critical roles of cognitive load and knowledge transfer in multi-domain learning scenarios, which are largely overlooked in existing single-domain KT methods.
    \item We propose a novel method, LT-MKT, for multi-domain knowledge tracing, which explicitly models cognitive load and knowledge transfer. In addition, LT-MKT leverages LLMs to construct a Multi-domain Hierarchical Graph for capturing semantic relationships across domains.
    \item Extensive experiments on four real-world datasets demonstrate that LT-MKT achieves state-of-the-art performance against 11 baselines, validating the effectiveness of modeling cognitive load and knowledge transfer in multi-domain learning scenarios.
\end{itemize}

\section{Related Work}
\subsection{Knowledge Tracing}
Knowledge Tracing (KT) is a fundamental task that aims to dynamically monitor and assess students' evolving knowledge states within online learning scenarios~\cite{shen2024KT,huang2019exploring,liu2019exploiting}. Over the years, a significant body of research has been dedicated to solving KT tasks. Early approaches primarily relied on statistical models, such as hidden Markov models~\cite{BKT, DBKT}, or logistic regression-based functions that incorporated various influencing factors~\cite{PFA, LFA, KTM}, to estimate students' conceptual mastery levels. With the rise of deep learning, Deep Knowledge Tracing (DKT) marked a pivotal moment by introducing neural networks into KT~\cite{DKT}, opening the door to more sophisticated, data-driven modeling techniques.
Following DKT, numerous variants of deep knowledge tracing have emerged, each introducing novel mechanisms to improve performance. These include memory-augmented methods~\cite{DKVMN,abdelrahman2019knowledge}, which explicitly model students' past interactions, graph-based approaches~\cite{gkt,tong2020hgkt,ni2023hhskt,zhang2022apgkt,wu2024graph} that better capture the relational structure between concepts, and attention-based models~\cite{ghosh2020AKT, SAKT, SAINT+}, which focus on highlighting key interactions during learning. Additionally, other KT models have sought to enhance student performance prediction by incorporating auxiliary information or designing more specialized network architectures~\cite{liu2019ekt,wang2022neuralcd,song2022bi,shen2021learning,wang2021temporal,lee2023difficulty,liu2024question,ma2024hd,sun2024interpretable,liu2024fdkt}.
Despite these advances, most existing KT methods mainly focus on single-domain learning scenarios, such as mathematics, while overlooking the influence of multi-domain learning behaviors. This domain-specific assumption limits the ability of these methods to exploit cross-domain knowledge dependencies, which may provide valuable contextual information for accurately understanding students' overall knowledge states.
Recently, several studies have begun exploring the use of multi-domain information to enhance KT performance. For instance, adaptive knowledge tracing methods~\cite{cheng2022adaptkt,tang2024domain,wu2025cross} leverage knowledge from related domains to alleviate data sparsity and improve performance in target domains. Meanwhile, some studies have started to investigate multi-domain knowledge tracing directly. PromptKT~\cite{liu2025prompt} introduces a prompt-enhanced paradigm that utilizes student interaction data from multiple domains to improve KT performance jointly across domains. TransKT~\cite{han2025contrastive} further proposes a contrastive cross-course knowledge tracing framework that leverages concept graph-guided knowledge transfer to model relationships among learning behaviors across different courses, thereby improving knowledge state estimation.
However, existing methods primarily focus on transferring information across domains while largely overlooking the underlying cognitive mechanisms that arise from multi-domain learning. In particular, they fail to explicitly model how cross-domain interactions contribute to cognitive load and knowledge transfer during the learning process, which implicitly influences students' knowledge states.
Different from previous studies, our work explicitly models the interplay of knowledge states across multiple domains from both cognitive load and knowledge transfer perspectives. By doing so, our method provides a more comprehensive representation of students' learning processes and better reflects the interconnected nature of real-world multi-domain learning scenarios.

\subsection{LLMs for Education}

Large Language Models (LLMs) have demonstrated remarkable effectiveness across a wide range of educational tasks, consistently achieving strong performance in learning-related applications~\cite{wang2024large,liu2026look,wang2026rec2,sha2026process,liu2026perma}. Recent studies show that LLMs can reach near student-level proficiency on standardized examinations in subjects such as mathematics and physics~\cite{achiam2023gpt}, highlighting their potential in educational scenarios including tutoring, writing assistance, and reading comprehension~\cite{malinka2023educational}. Moreover, LLMs have shown strong capabilities in enabling personalized learning and supporting automated educational assessment~\cite{li2024learning,liu2024socraticlm,wang2026personalized,lv2025genal}.

In the domain of Knowledge Tracing (KT), recent research has increasingly explored the use of LLMs to enhance representation learning and semantic understanding from both student interactions and textual educational resources. For example, LLM-SBCL~\cite{ni2024enhancing} utilizes LLMs to analyze the textual content of questions within student-question interaction networks, enabling more accurate identification of underlying knowledge concepts, especially in cold-start scenarios with limited data. Similarly, DCL4KT+LLM~\cite{lee2023difficulty} leverages LLMs to estimate question difficulty from question stems and associated knowledge concepts, effectively alleviating the issue of missing difficulty annotations for unseen questions. In addition,~\citet{sonkar2023deduction} explores the reasoning capabilities of LLMs to simulate students' misconceptions and incorrect responses based on detailed knowledge profiles, demonstrating the potential of LLMs for modeling complex learning behaviors.
More recently, several studies have begun integrating LLMs into knowledge structure modeling for KT. SINKT~\cite{fu2024sinkt} introduces the first inductive knowledge tracing framework that directly incorporates LLMs to enhance representation learning and improve generalization to unseen questions and students. TransKT~\cite{han2025contrastive} further employs concept graph-guided knowledge transfer to model relationships among learning behaviors across different courses, demonstrating the effectiveness of graph-based knowledge structures in capturing semantic dependencies across domains. These studies collectively demonstrate the strong capability of LLMs in extracting semantic relationships from educational content and constructing meaningful knowledge representations for KT tasks.
Different from previous studies that mainly employ LLMs to model relationships among concepts, we leverage LLMs to construct a multi-domain hierarchical graph that jointly captures question-concept and concept-concept relationships across domains, providing the structural foundation for modeling cognitive load and knowledge transfer in multi-domain knowledge tracing.

\section{Problem Definition}

In an Intelligent Tutoring System (ITS), we consider a set of students $\mathcal{S}$, a set of questions $\mathcal{Q}$, and a set of knowledge concepts $\mathcal{C}$ distributed across multiple domains $\mathcal{D}$. For a student $s \in \mathcal{S}$, the learning history is represented as:
\begin{equation}
    R_s = \{(q_1^{d_1}, r_1), (q_2^{d_2}, r_2), \dots, (q_T^{d_k}, r_T)\},
\end{equation}
where $q_t^{d_j} \in \mathcal{Q}$ denotes the question answered by the student at time step $t$, which belongs to domain $d_j \in \mathcal{D}$, and $r_t \in \{0,1\}$ represents the student's response correctness, where $r_t = 1$ indicates a correct response and $r_t = 0$ otherwise.
The goal of multi-domain knowledge tracing is to predict the probability that the student correctly answers the next question $q_{T+1}^{d_m}$ from domain $d_m \in \mathcal{D}$, based on the historical interaction sequence $R_s$. Formally, the prediction objective is defined as:
\begin{equation}
    p(r_{T+1} = 1 \mid R_s, q_{T+1}^{d_m}).    
\end{equation}

Different from traditional single-domain knowledge tracing, where all learning interactions are restricted to a single domain, multi-domain knowledge tracing involves more complex cross-domain learning behaviors. Specifically, students may frequently switch across different domains during learning, while individual questions may simultaneously require knowledge from multiple domains, thereby introducing additional factors that complicate knowledge state modeling and make it more challenging.

\section{The LT-MKT Model}
\begin{figure*}[t]
  \centering
  \includegraphics[width=0.98\linewidth]{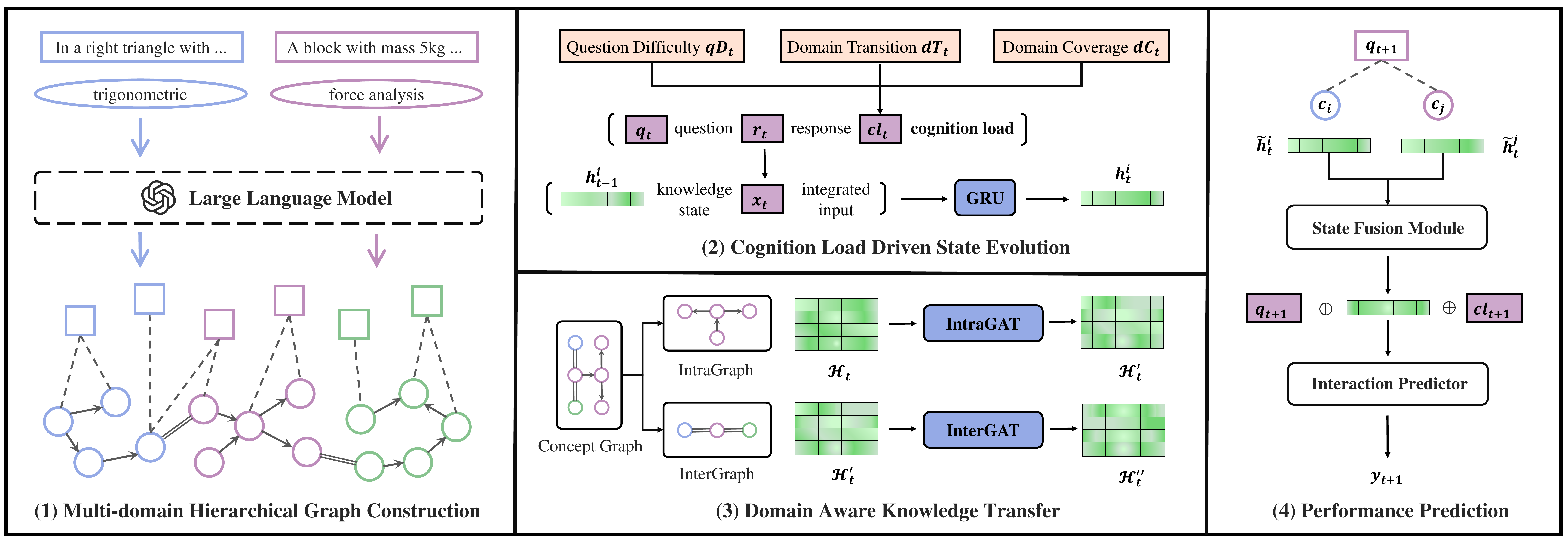}
  \caption{Overall framework of LT-MKT. The model first constructs a multi-domain hierarchical graph from question texts and knowledge concepts with LLM-guided reasoning. Based on this graph, LT-MKT models cognitive load through question difficulty, domain transition, and domain coverage, and updates students' knowledge states via GRU-based state evolution. It then performs domain-aware knowledge transfer with intra-domain prerequisite propagation and inter-domain correlation propagation, followed by state fusion and performance prediction for the next interaction.}
  \label{fig:model}
   \vspace{-0.2cm}
\end{figure*}

In this section, we present the proposed LT-MKT model in detail. An overview of the overall framework is illustrated in Figure~\ref{fig:model}. Specifically, we first introduce the Multi-domain Hierarchical Graph, which serves as the relational foundation for modeling semantic dependencies across domains (\S~\ref{sec: graph}). Based on this graph structure, we then model cognitive load (\S~\ref{sec:cog}) and knowledge transfer (\S~\ref{sec:tans}) to refine students' knowledge state estimation in multi-domain learning scenarios, ultimately enabling more accurate student performance prediction (\S~\ref{sec:pred}).

\subsection{Hierarchical Graph Construction}\label{sec: graph}

In multi-domain learning scenarios, knowledge concepts are often interconnected across domains, making it difficult to capture the semantic dependencies required for modeling cognitive load and knowledge transfer. To address this issue, we construct a Multi-domain Hierarchical Graph (MDHG) to provide a unified relational structure for multi-domain knowledge tracing.

Formally, let $\mathcal{G}=(\mathcal{Q},\mathcal{C},\mathcal{E})$ denote the MDHG, where $\mathcal{Q}$ is the question set, $\mathcal{C}$ is the knowledge concept (KC) set, and $\mathcal{E}$ contains three types of edges: question-to-concept \textit{associations}, intra-domain concept \textit{prerequisites}, and inter-domain concept \textit{correlations}. Association edges indicate the concepts involved in each question. Prerequisite edges describe directed learning dependencies within the same domain, while correlation edges capture semantic or functional connections across domains.

Constructing such educational graphs usually requires extensive expert annotation. To improve scalability, we employ LLMs to construct the MDHG through prompt-guided reasoning. Specifically, for each question, the prompt provides the question text, domain information, main concept, candidate concepts, and existing prerequisite relations. Based on chain-of-thought reasoning~\cite{wei2022chain}, the LLM first identifies the concepts involved in the question, then selects prerequisite concepts from the same domain and correlated concepts from other domains. For a question $q_j$ with primary concept $c_i$, this process is formulated as:
\begin{equation}
    \mathcal{A}(q_j), \mathcal{M}(q_j), \mathcal{N}(q_j)
    = \mathrm{LLM}_{\theta}(\mathcal{P}(q_j,d_j,c_i,\mathcal{C},\mathcal{E}_{pre})),
\end{equation}
where $\mathcal{A}(q_j)$ denotes the associated concepts of $q_j$, $\mathcal{M}(q_j)$ denotes intra-domain prerequisite concepts, $\mathcal{N}(q_j)$ denotes inter-domain correlated concepts, and $\mathcal{P}(\cdot)$ is the graph construction prompt.

After processing all questions, we aggregate the generated relations at the concept level. For each concept $c_i$, let $\mathcal{Q}_i$ be the set of questions whose primary concept is $c_i$. The prerequisite and correlation sets of $c_i$ are obtained by:
\begin{equation}
    \mathcal{M}_i = \bigcup_{q_j \in \mathcal{Q}_i} \mathcal{M}(q_j), \quad
    \mathcal{N}_i = \bigcup_{q_j \in \mathcal{Q}_i} \mathcal{N}(q_j).
\end{equation}
We then construct the final edge sets as:
\begin{equation}
\begin{aligned}
    \mathcal{E}_{qc} &= \{(q_j,c) \mid c \in \mathcal{A}(q_j)\},\\
    \mathcal{E}_{pre} &= \{(c_m,c_i) \mid c_m \in \mathcal{M}_i\},\\
    \mathcal{E}_{cor} &= \{(c_i,c_n) \mid c_n \in \mathcal{N}_i\}.
\end{aligned}
\end{equation}
Thus, the complete edge set is $\mathcal{E}=\mathcal{E}_{qc}\cup\mathcal{E}_{pre}\cup\mathcal{E}_{cor}$.

To ensure educational consistency, prerequisite relations are constrained to satisfy the DAG property:
\begin{equation}
    \mathrm{Cycle}(\mathcal{C},\mathcal{E}_{pre})=\varnothing,
\end{equation}
which avoids circular dependencies in learning paths. Since each new concept can be processed with the same prompt and aggregation procedure, the graph can also be incrementally extended when unseen concepts appear in future learning scenarios. 
Detailed prompts are provided in the code repository.

\subsection{Cognitive Load Driven State Evolution}\label{sec:cog}

Cognitive load theory~\cite{plass2010cognitive} posits that cognitive load stems from the \textit{complexity of learning materials and interactions}, thereby implicitly influencing students’ knowledge acquisition and the evolution of their knowledge states during the learning process. In multi-domain learning scenarios, cognitive load is further amplified by cross-domain learning behaviors in both temporal and knowledge dimensions. Therefore, accurately modeling students' knowledge states requires explicitly capturing the dynamics of cognitive load throughout the learning process.
Motivated by findings from existing studies~\cite{lee2023difficulty,wu2025cross}, we investigate the effects of three key factors related to cognitive load: \textbf{question difficulty, domain transition}, and \textbf{domain coverage}. These factors characterize different aspects of cognitive complexity introduced by multi-domain learning interactions. 
While acknowledging the potential influence of other factors, we leave their exploration for future work.

\paragraph{Question Difficulty}  
Question difficulty reflects the \textit{complexity of learning materials} to some extent and serves as a critical factor in assessing cognitive load. It has been proven to be a critical factor in accurately estimating students' knowledge states~\cite{lee2023difficulty}.
Following some previous works~\cite{shen2022assessing}, we calculate the difficulty of the question $q_i$ as follows:
\begin{equation}\label{eq:diff}
    qD_i = \frac{\sum_{i=1}^{|S_i|}a_i == 0}{|S_i|}\times\lambda_P,
\end{equation}
where $S_i$ denotes the set of students who have attempted question $q_i$, and $a_{s, i} \in \{0,1\}$ indicates whether student $s$ answers question $q_i$ correctly (1 for correct and 0 for incorrect). $\lambda_P$ represents the predefined granularity level of question difficulty. Intuitively, a question is considered more difficult if it is answered incorrectly by a larger proportion of students.
We further represent the difficulty embedding using an embedding matrix $\mathbf{E}_{qD} \in \mathbb{R}^{\lambda_P \times d_{qD}}$, where $d_{qD}$ denotes the embedding dimension. 
\paragraph{Domain Transition} \label{sec:dt}  
Domain transition refers to shifts between domains during learning, reflecting the \textit{complexity of interactions} that influence cognitive load and subsequently affect knowledge state estimation. 
Intuitively, given a student’s interaction history $\mathcal{R}_s = \{(q_1^{d_1}, r_1), (q_2^{d_2}, r_2), \dots, (q_T^{d_k}, r_T)\}$, the domain transition factor up to time $t$ can be calculated as:
\begin{equation}\label{eq:dt}
{dT}_t = \sum_{t'= t-ws}^t (d_{t'} \neq d_{t'-1}),
\end{equation}
where $ws$ is the size of the time window. 
Since a student's learning is more strongly influenced by recent interactions, considering transitions over the entire time span may not be appropriate. 
In practice, a student who frequently switches between different domains within a short period will have a higher domain transition value, indicating stronger cross-domain cognitive switching effects.
Then we represent the transitions embedding with embedding matrix $\textbf{E}_{dT} \in \mathbb{R}^{\lambda_T\times d_{dT}}$, where $\lambda_T$ represents the maximum value of domain transitions observed across all sequences, $d_{dT}$ is the dimension.
\paragraph{Domain Coverage} 
Domain coverage refers to the number of domains covered by the concepts assessed in a specific question. Intuitively, the more domains a question involves, the higher its \textit{complexity} and the greater the \textit{interaction} required for learning. This increased complexity contributes to higher cognitive load, which in turn affects knowledge state estimation.
For a question $q_i$, the domain coverage is defined as follows:
\begin{equation}
    \text dC_i = |D_i|,
\end{equation}
where $\mathcal{D}_i = \{d_1, d_2,\dots\}$ represents its relevant concepts coverage couple of domains. 
Similarly, the domain coverage is represented as an embedding matrix $\textbf{E}_{dC} \in \mathbb{R}^{\lambda_C \times d_{dC}}$, where $\lambda_C$ represents the total number of domains and $d_{dC}$ is the dimension.

\paragraph{State Evolution}
After examining the impact of cognitive load through the three key factors discussed above, we define the cognitive load $cl_t$ at time $t$ as:
\begin{equation}\label{eq:clt}
    \textbf{cl}_t = \textbf{qD}_t \oplus \textbf{dT}_t \oplus \textbf{dC}_t.
\end{equation}

Subsequently, we integrate the cognitive load into the interaction embedding $\textbf{x}_t$ as follows:
\begin{equation}
    \textbf{x}_t = \textbf{q}_t \oplus \textbf{r}_t \oplus \textbf{cl}_t,
\end{equation}
where $\textbf{q}_t$ is obtained by encoding the question text using BERT~\cite{devlin2019bert}, and for the answer $\textbf{r}_t$, i.e., 0 or 1, we expand it to an all-zero or all-one vector $\textbf{r}_t \in \mathbb{R}^{d_a}$, $d_a$ is the dimension.

Finally, we use Gated Recurrent Unit (GRU) \cite{chung2014empirical} to update the knowledge state to model the temporal effect in the learning process:
\begin{align}
    \mathbf{r}_t &= \sigma (\mathbf{W}_r \left[\mathbf{h}_{t-1} \oplus \mathbf{x}_t\right] + \mathbf{b}_r), \\
    \mathbf{z}_t &= \sigma (\mathbf{W}_z \left[\mathbf{h}_{t-1} \oplus \mathbf{x}_t\right] + \mathbf{b}_z), \\
    \mathbf{\tilde{h}}_{t} &= \tanh (\mathbf{W}_{\tilde{h}} \left[\mathbf{r}_t \cdot \mathbf{h}_{t-1} \oplus \mathbf{x}_t\right] + \mathbf{b}_{\tilde{h}}),\\
    \mathbf{h}_{t} &= (1 - \mathbf{z}_t) \cdot \mathbf{\tilde{h}}_{t} + \mathbf{z}_t \cdot \mathbf{h}_{t-1},
\end{align}
where $\mathbf{W}_r, \mathbf{W}_z, \mathbf{W}_{\tilde{h}} \in \mathbb{R}^{(d_k+d_x) \times d_k}$ and $\mathbf{b}_r, \mathbf{b}_z, \mathbf{b}_{\tilde{h}} \in \mathbb{R}^{d_k}$ and $\sigma$ denotes the \textit{sigmoid} function. $\mathbf{h}_t \in \mathbb{R}^{d_k}$ is a row extracted from the knowledge state matrix $\mathcal{H}_t$, which represents the knowledge state of the primary KC of the current question.

\subsection{Domain Aware Knowledge Transfer}\label{sec:tans}
Transfer of learning theory~\cite{cormier2014transfer} emphasizes that knowledge concepts are inherently interconnected, and knowledge transfer occurs when understanding one concept influences the learning and understanding of related concepts. Different from traditional single-domain knowledge tracing, where all concepts are confined to a single domain, multi-domain learning requires modeling knowledge propagation both within and across domains. 
Specifically, knowledge transfer in multi-domain learning can be categorized into two types: \textbf{vertical transfer} and \textbf{lateral transfer}~\cite{cormier2014transfer}. Vertical transfer refers to prerequisite-driven knowledge propagation within the same domain, where mastery of foundational concepts is essential for learning subsequent concepts. For instance, learning addition can facilitate understanding of multiplication within mathematics. In contrast, lateral transfer captures semantic or functional associations across different domains, where knowledge from one domain may indirectly influence related knowledge states in another domain. For instance, knowledge of trigonometric functions in mathematics can support the learning of force decomposition in physics.
Furthermore, learning hierarchy theory~\cite{khan2006hierarchy} posits that knowledge transfer within and across domains may occur at different stages of learning. In particular, intra-domain knowledge transfer is typically more immediate, as prerequisite dependencies directly affect subsequent learning within the same domain before cross-domain transfer takes effect~\cite{macaulay1999transfer}. 
Motivated by these inspirations, we design two graph attention modules, namely \textbf{intraGAT} and \textbf{interGAT}, to successively model intra-domain and inter-domain knowledge transfer, respectively. Specifically, \textbf{intraGAT} captures prerequisite-based knowledge propagation within the same domain, while \textbf{interGAT} further models cross-domain semantic interactions among related concepts.

To be specific, we first decompose the MDHG into two sub-graphs: intra-graph and inter-graph, as follows:
\begin{equation}
    \boldsymbol{\mathcal{G} = \mathcal{G}^{intra} + \mathcal{G}^{inter}},
\end{equation}
where $\mathcal{G}$, $\mathcal{G}^{intra}$ and $\mathcal{G}^{inter}$ are relation matrices. 

Then we design the intraGAT layer to capture within-domain knowledge transfer in $\mathcal{G}^{intra}$:
\begin{equation}
    \alpha^{intra}_{ij} = 
\frac{
\exp\left(\text{LeakyReLU}\left(a^T \left[ \mathbf{W} h_i \oplus \mathbf{W} h_j \right]\right)\right)
}{
\sum_{k \in \mathcal{M}_i} \exp\left(\text{LeakyReLU}\left(a^T \left[ \mathbf{W} h_i \oplus \mathbf{W} h_k \right]\right)\right)
},
\end{equation}
where $\mathbf{a} \in \mathbb{R}^{2d_k}$ is a learnable weight vector, and LeakyReLU denotes the activation function with a negative slope coefficient $\alpha = 0.2$. $\mathcal{M}_i$ represents the set of intra-domain successor KCs associated with knowledge concept $c_i$. Furthermore, to maintain the stability of students' knowledge states and avoid unnecessary propagation noise, we only update the primary knowledge concept $c_i$ and its associated concepts in $\mathcal{M}_i$ according to the question-to-concept associations in the MDHG:
\begin{equation}
    h'_i = \text{ELU}\left( \sum_{j\in\mathcal{M}_i}\alpha^{intra}_{ij}\textbf{W} h_j\right),
\end{equation}
where ELU is the exponential linear units activation function. Similarly, the interGAT layer is designed to capture knowledge transfer across domains:
\begin{equation}
    \alpha^{inter}_{ij} = 
\frac{
\exp\left(\text{LeakyReLU}\left(a^T \left[ \mathbf{W} h'_i \oplus \mathbf{W} h'_j \right]\right)\right)
}{
\sum_{k \in \mathcal{N}_i} \exp\left(\text{LeakyReLU}\left(a^T \left[ \mathbf{W} h'_i \oplus \mathbf{W} h'_k \right]\right)\right)
},
\end{equation}
\begin{equation}
    h''_i = \text{ELU}\left( \sum_{j\in\mathcal{N}_i}\alpha^{inter}_{ij}\textbf{W} h'_j\right),
\end{equation}
where $\mathcal{N}_i$ is the set of inter-domain correlated KCs of KC $c_i$. After knowledge transfer within and across domains, we finally get the student's final knowledge state $\mathcal{H}^{''}_{t}$.

\subsection{Prediction and Objective Function}\label{sec:pred}
After modeling cognitive load and knowledge transfer across domains, we obtain the student's updated knowledge state representation $\mathcal{H}^{''}_{t}$. In LT-MKT, the prediction of the student's performance on the next question $q_{t+1}$ is based on three components: the question embedding $\boldsymbol{q}_{t+1}$, the cognitive load representation $\boldsymbol{cl}_{t+1}$, and the fused knowledge state vector $\boldsymbol{\tilde{h}}_{t}$. Specifically, the fused knowledge state is computed through a State Fusion Module as:
\begin{align}
    \boldsymbol{\tilde{h}}_{t} &= F(\boldsymbol{\tilde{h}}^i_{t} + \boldsymbol{\tilde{h}}^j_{t} + \cdots),
\end{align}
where $F(\cdot)$ denotes a mean fusion operation, which has been widely adopted as an effective strategy for aggregating multiple representations. $\boldsymbol{\tilde{h}}^i_{t}$ represents the knowledge state corresponding to concept $c_i$ extracted from $\mathcal{H}^{''}_{t}$, while $\boldsymbol{\tilde{h}}^j_{t}$ denotes the knowledge state corresponding to another related concept $c_j$ involved in question $q_{t+1}$. Other concept-specific knowledge states are obtained similarly.
The predicted probability of correctly answering question $q_{t+1}$ is then computed as:
\begin{align}
    y_{t+1} &= \sigma(\mathbf{W}_p[\boldsymbol{q}_{t+1} \oplus \boldsymbol{\tilde{h}}_{t} \oplus \boldsymbol{cl}_{t+1}] + \mathbf{b}_p),
\end{align}
where $\mathbf{W}_p \in \mathbb{R}^{(d_q + d_k + d_{cl}) \times d_k}$ and $\mathbf{b}_p \in \mathbb{R}^{d_k}$ are trainable parameters, $\oplus$ denotes the concatenation operation, and $\sigma(\cdot)$ represents the sigmoid activation function. The output $y_{t+1} \in (0,1)$ denotes the probability that the student correctly answers question $q_{t+1}$.

To optimize LT-MKT, we adopt the binary cross-entropy loss between the predicted response $y_{t}$ and the ground-truth response $r_{t}$ as the training objective:
\begin{equation}
    \mathcal{L} = - \sum_{t=1}^{T} \left( r_t \log y_t + (1 - r_t) \log (1 - y_t) \right).
\end{equation}

Minimizing this objective encourages the model to accurately estimate students' future performance by jointly modeling cognitive load and knowledge transfer in multi-domain learning scenarios.

\begin{table}[]
\centering
\caption{Dataset statistics}
\setlength{\tabcolsep}{4pt}
\begin{tabular}{lllll}
\hline
Datasets       & JuniorH & SeniorH & PTADiscJP & PTADiscDS  \\ 
\hline
\#Students     & 1,081        &4,869     & 29,430       & 12,271   \\
\#Concepts     & 139         &198      & 1,245       & 634   \\
\#Questions    & 170         &264      & 27,820       & 18,702 \\
\#Interactions & 39,230       &133,683   & 11,172,165       & 1,788,245 \\
\#Avg.Inter    & 36.29       &27.45    &  379.6      & 145.73   \\ 
\#Avg.Cross    & 14.26       &12.39    &  8.72      & 7.81    \\ 
\hline
\end{tabular}
\label{tab:datasta}
  \vspace{-0.3cm}
\end{table}

\begin{table*}[t]
\centering
\caption{Results of all comparison methods on the student performance prediction task. Existing state-of-the-art results are marked by the underline, and the best results are bold. * indicates p-value \textless 0.05 in the t-test.}
\label{tab:results}
\begin{tabular}{lcccccccccccc}
\toprule
\multirow{2}{*}{Methods} & \multicolumn{3}{c}{JuniorH} & \multicolumn{3}{c}{SeniorH} & \multicolumn{3}{c}{PTADiscJP} & \multicolumn{3}{c}{PTADiscDS} \\
\cmidrule(lr){2-4} \cmidrule(lr){5-7} \cmidrule(lr){8-10} \cmidrule(lr){11-13}
& AUC$~\uparrow$ & ACC$~\uparrow$ & RMSE$~\downarrow$ & AUC$~\uparrow$ & ACC$~\uparrow$ & RMSE$~\downarrow$ & AUC$~\uparrow$ & ACC$~\uparrow$ & RMSE$~\downarrow$ & AUC$~\uparrow$ & ACC$~\uparrow$ & RMSE$~\downarrow$ \\ 
\midrule
DKT & 0.8840 & 0.7991 & 0.3658 & 0.8861 & 0.7926 & 0.3711 & 0.7246 & 0.7843 & 0.3687 & 0.6504 & 0.7602 & 0.4358 \\
GKT & 0.8818 & 0.7983 & 0.3639 & 0.8883 & 0.7828 & 0.3730 & 0.7298 & 0.7892 & 0.3664 & 0.6534 & 0.7646 & 0.4335 \\
AKT & 0.8871 & 0.8032 & 0.3662 & 0.8864 & 0.7928 & 0.3718 & 0.7307 & 0.7904 & 0.3655 & 0.6543 & 0.7651 & 0.4337 \\
HawkesKT & 0.7515 & 0.7202 & 0.4768 & 0.7610 & 0.7124 & 0.4874 & 0.7441 & 0.7903 & 0.3612 & 0.6574 & 0.7641 & 0.4358 \\
LPKT & 0.8844 & 0.8004 & 0.3654 & 0.8850 & 0.7922 & 0.3719 & 0.7274 & 0.7876 & 0.3669 & 0.6575 & 0.7634 & 0.4346 \\
DIMKT & 0.8895 & 0.8035 & 0.3632 & 0.8865 & 0.7927 & 0.3707 & 0.7346 & 0.7948 & 0.3547 & 0.6573 & 0.7701 & 0.4328 \\
AT-DKT & 0.8855 & 0.8018 & 0.3652 & 0.8851 & 0.7898 & 0.3722 & 0.7285 & 0.7884 & 0.3664 & 0.6619 & 0.7628 & 0.4351 \\
MIKT & 0.8845 & 0.7969 & 0.3680 & 0.8941 & 0.7991 & 0.3640 & 0.7382 & 0.8006 & 0.3577 & 0.6646 & 0.7741 & 0.4316 \\
SINKT & 0.8947 & 0.8114 & 0.3603 & 0.8986 & 0.8075 & 0.3614 & 0.7473 & 0.8118 & 0.3532 & 0.6701 & 0.7842 & 0.4286 \\
promptKT & 0.8981 & 0.8142 & 0.3586 & 0.9013 & 0.8115 & 0.3597 & 0.7407 & 0.8141 & 0.3611 & 0.6682 & 0.7871 & 0.4276 \\
TransKT & \underline{0.9143} & \underline{0.8272} & \underline{0.3544} & \underline{0.9174} & \underline{0.8241} & \underline{0.3567} & \underline{0.7514} & \underline{0.8213} & \underline{0.3517} & \underline{0.6802} & \underline{0.7943} & \underline{0.4148} \\
\textbf{LT-MKT} & \textbf{0.9387*} & \textbf{0.8425*} & \textbf{0.3484*} & \textbf{0.9312*} & \textbf{0.8470*} & \textbf{0.3378*} & \textbf{0.7645*} & \textbf{0.8410*} & \textbf{0.3485*} & \textbf{0.6929*} & \textbf{0.8031*} & \textbf{0.4065*} \\
\bottomrule
\end{tabular}
\end{table*}

\section{Experiments}
In this section, we first introduce the datasets, followed by a description of the baseline models and training details. Subsequently, we present the results of extensive experiments. 

\subsection{Datasets}
We conduct experiments on four real-world multi-domain learning datasets. Two publicly available datasets are derived from the PTADisc dataset~\cite{hu2023ptadisc}~\footnote{https://github.com/wahr0411/PTADisc}, namely \textbf{PTADiscJP} and \textbf{PTADiscDS}. Specifically, PTADiscJP contains learning records from two programming-related domains: Java and Python (Java\&Python), while PTADiscDS consists of records from C programming and Data Structure \& Algorithm Analysis (C\&DS).  
In addition, we conduct experiments on two proprietary datasets supplied by iFLYTEK Co., Ltd., collected from the intelligent learning machine~\footnote{https://xxj.xunfei.cn/}: \textbf{JuniorH} and \textbf{SeniorH}. These datasets contain students' question-answering records across three academic domains: mathematics, physics, and English. Specifically, the \textbf{JuniorH} dataset includes learning records from grades 7--9, while the \textbf{SeniorH} dataset contains records from grades 10--12.
Basic statistics of all datasets are summarized in Table~\ref{tab:datasta}.

It is worth noting that, as shown in Table~\ref{tab:datasta}, we calculate the average number of Domain Transitions (\#Avg.Cross) for students in real-world learning scenarios (detailed in Section~\ref{sec:dt}, where $ws$ is set equal to $t$). The results indicate that students frequently engage in cross-domain learning in multi-domain educational settings.

\subsection{Baselines}
To validate the effectiveness of our proposed LT-MKT in multi-domain learning scenarios, we selected eleven representative KT models as baselines. Their details are as follows:
\begin{itemize}[leftmargin=*,itemsep=0.8pt]
    \item \textbf{DKT}~\cite{DKT} utilizes RNN to model the learning sequence, where the hidden state represents the learner's knowledge state.
    \item \textbf{GKT}~\cite{gkt} generates a transition graph from the dataset and employs GNN to encode students' knowledge states.
    \item \textbf{AKT}~\cite{ghosh2020AKT} uses a monotonic attention mechanism to capture dependencies in learning sequences. 
    \item \textbf{Hawkes-KT}~\cite{wang2021temporal} models temporal cross-effects from point process, where prior interactions impact skill mastery over time.
    \item \textbf{LPKT}~\cite{shen2021learning} models students' progress by considering interval times to calculate learning gains and forgetting rates.
    \item \textbf{DIMKT}~\cite{shen2022assessing} is a sequential model incorporating question/skill difficulty levels as inputs.
    \item \textbf{AT-DKT}~\cite{chen2023improving} enhances DKT by including two auxiliary tasks: question tagging and predicting students’ prior knowledge.
    \item \textbf{MIKT}~\cite{sun2024interpretable} models students' knowledge states at both coarse-grained domain and fine-grained concept levels.
    \item \textbf{SINKT}~\cite{fu2024sinkt} is a structure-aware inductive model that harnesses large language models to effectively generalize to new student responses and unseen questions.
    \item \textbf{promptKT}~\cite{liu2025prompt} introduces a prompt-enhanced paradigm utilizing a pre-trained Transformer backbone and a soft domain prompt module for multi-domain knowledge tracing.
    \item \textbf{TransKT}~\cite{han2025contrastive} leverages concept graph guided knowledge transfer to model the relationships between learning behaviors across different courses.
\end{itemize}

\subsection{Experimental Setup}
In our experiments, we split the dataset in an 8:1:1 ratio by learners to obtain the training set, validation set, and testing set. The difficulty granularity parameter $\lambda_P$ in Eq.~\ref{eq:diff} is set to 30, and the sliding window size $ws$ in Eq.~\ref{eq:dt} is set to 20. 
During graph construction, we employed Qwen-plus~\footnote{https://bailian.console.aliyun.com/} (default) as the LLM backbone to balance capability and cost. The sampling temperature is set to 0 to reduce randomness.
All models are optimized using the Adam optimizer. The learning rate is selected from $\{0.001, 0.005\}$ based on validation performance, with a learning rate decay applied every 10 epochs. Hyperparameters are tuned on the validation set, and the model achieving the best validation performance is used for final evaluation.
Following previous KT studies, we adopt three widely used evaluation metrics to comprehensively assess model performance: Area Under the ROC Curve (AUC), Accuracy (ACC), and Root Mean Squared Error (RMSE). All experiments were conducted on a cluster of Linux servers with Tesla V100 GPUs.

\begin{table*}[t]
\centering
\caption{Results of ablation experiments across four datasets.}
\label{tab:ablation_study}
\begin{tabular}{lcccccccccccc}
\toprule
\multirow{2}{*}{Methods} & \multicolumn{3}{c}{JuniorH} & \multicolumn{3}{c}{SeniorH} & \multicolumn{3}{c}{PTADiscJP} & \multicolumn{3}{c}{PTADiscDS} \\
\cmidrule(lr){2-4} \cmidrule(lr){5-7} \cmidrule(lr){8-10} \cmidrule(lr){11-13}
& AUC$~\uparrow$ & ACC$~\uparrow$ & RMSE$~\downarrow$ & AUC$~\uparrow$ & ACC$~\uparrow$ & RMSE$~\downarrow$ & AUC$~\uparrow$ & ACC$~\uparrow$ & RMSE$~\downarrow$ & AUC$~\uparrow$ & ACC$~\uparrow$ & RMSE$~\downarrow$ \\ 
\midrule
w/o CL & 0.9005 & 0.8116 & 0.3636 & 0.8935 & 0.8150 & 0.3561 & 0.7471 & 0.8237 & 0.3610 & 0.6673 & 0.7896 & 0.4231 \\
w/o TG & 0.9227 & 0.8322 & 0.3491 & 0.9160 & 0.8360 & 0.3536 & 0.7594 & 0.8302 & 0.3544 & 0.6799 & 0.7955 & 0.4102 \\
w/o SF & 0.9320 & 0.8419 & 0.3503 & 0.9252 & 0.8455 & 0.3394 & 0.7603 & 0.8388 & 0.3505 & 0.6865 & 0.8011 & 0.4097 \\
\textbf{LT-MKT} & \textbf{0.9387} & \textbf{0.8425} & \textbf{0.3484} & \textbf{0.9312} & \textbf{0.8470} & \textbf{0.3378} & \textbf{0.7645} & \textbf{0.8410} & \textbf{0.3485} & \textbf{0.6929} & \textbf{0.8031} & \textbf{0.4065} \\
\bottomrule
\end{tabular}
\end{table*}

\subsection{Overall Performance}
We compare LT-MKT with eleven representative baselines, and the results are reported in Table~\ref{tab:results}. Several observations can be drawn from the results.
First, LT-MKT consistently achieves the best performance across all datasets and evaluation metrics, demonstrating the effectiveness of explicitly modeling cognitive load and knowledge transfer in multi-domain knowledge tracing.
Second, LT-MKT achieves more significant improvements on JuniorH and SeniorH, likely because the denser cross-domain learning behaviors (\#Avg.Cross in Table~\ref{tab:datasta}) further validate the rationality and superiority of our method.
Third, we have noticed that HawkesKT performs noticeably worse than other methods, especially on JuniorH and SeniorH. This may result from its sensitivity to temporal dynamics, while the relatively short interaction sequences in these datasets (Table~\ref{tab:datasta}) limit its effectiveness.
Finally, LLM-based KT methods (e.g., SINKT, TransKT) generally outperform traditional KT models, indicating that LLMs are effective at capturing richer semantic relationships for knowledge tracing.

\subsection{Ablation Study}
To further investigate the importance of each module in LT-MKT, we design three variations to conduct the ablation study, each of which removes one part from the original method:
\begin{itemize}[leftmargin=*,itemsep=0.8pt]
    \item \textbf{LT-MKT w/o CL}, which removes cognitive load effects including question difficulty, domain transition, and domain coverage features (section~\ref{sec:cog}). 
    \item \textbf{LT-MKT w/o TG}, replaces the intraGAT and interGAT layers with a single GAT layer (section~\ref{sec:tans}).
    \item \textbf{LT-MKT w/o SF}, which excludes the state fusion module $F$ in the prediction stage (section~\ref{sec:pred}).
\end{itemize}

From Table~\ref{tab:ablation_study}, several key findings can be drawn. Firstly, the complete model achieved the best overall performance. 
Secondly, the cognitive load effect significantly impacts model performance, emphasizing the importance of cross-domain features in multi-domain learning.
Thirdly, knowledge transfer mainly affects cross-domain knowledge evolution, and its removal causes a performance drop, while state fusion has a smaller effect.
The collaborative synergy of the components leads to optimal results, as the absence of any component results in a decline in performance.

\subsection{Effect of the Graph Construction}
Since LT-MKT relies on LLMs to construct the Multi-domain Hierarchical Graph (MDHG), we perform a human evaluation to assess the \textit{quality of graphs} generated by \textit{different LLM backbones} from both educational and structural perspectives.
Specifically, we recruit twenty experts for evaluation, including five students with relevant coursework experience, eight researchers in educational data mining, and seven teachers with practical experience in curriculum design and assessment. We compare seven representative LLM backbones with different scales and families, including Qwen2.5-7B, Qwen2.5-32B, Qwen2.5-72B, Llama3-70B, DeepSeek-R1, Claude3 Sonnet, and GPT-5 pro.
Each generated graph is rated on two dimensions: (i) \textit{Educational Rationality (ER)}, which measures whether the prerequisite and correlation relations are pedagogically meaningful and consistent with real learning dependencies; and (ii) \textit{Structural Consistency (SC)}, which evaluates whether the graph exhibits coherent structure without invalid or circular prerequisite relations. Both metrics are scored on a five-point Likert scale (1--5). We further report Fleiss' $\kappa$~\cite{fleiss1971measuring} to measure inter-rater agreement.

The results are shown in Table~\ref{tab:graph_quality}. Among all compared models, GPT-5 achieves the best performance across all metrics, including the highest inter-annotator agreement (Fleiss’ $\kappa$). Overall, larger and more capable models consistently achieve higher ER and SC scores, indicating stronger semantic understanding and structural reasoning ability in graph construction. These trends further highlight the compatibility and extensibility of our framework, as LT-MKT consistently benefits from stronger LLM backbones.
With the release of more powerful models in the future, the quality of graph construction is expected to be further improved, thereby providing an even stronger foundation for multi-domain knowledge tracing.

\begin{table}[t]
\centering
\caption{Expert evaluation results of graph quality generated by different LLM backbones.}
\label{tab:graph_quality}
\begin{tabular}{lccc}
\toprule
LLM Backbone & ER $\uparrow$ & SC $\uparrow$ & Fleiss' $\kappa$ $\uparrow$ \\
\midrule
Qwen2.5-7B   & 3.72 & 3.60 & 0.61 \\
Qwen2.5-32B  & 4.07 & 3.95 & 0.69 \\
Qwen2.5-72B  & 4.29 & 4.16 & 0.74 \\
Llama3-70B   & 4.11 & 4.02 & 0.71 \\
DeepSeek-R1    & 4.13 & 3.98 & 0.68 \\
Claude3 Sonnet       & 4.45 & 4.38 & 0.79 \\
\textbf{GPT-5 pro}        & \textbf{4.66} & \textbf{4.56} & \textbf{0.84} \\
\bottomrule
\end{tabular}
  \vspace{-0.3cm}
\end{table}

\begin{figure}[t]
  \centering
  \includegraphics[width=0.98\linewidth]{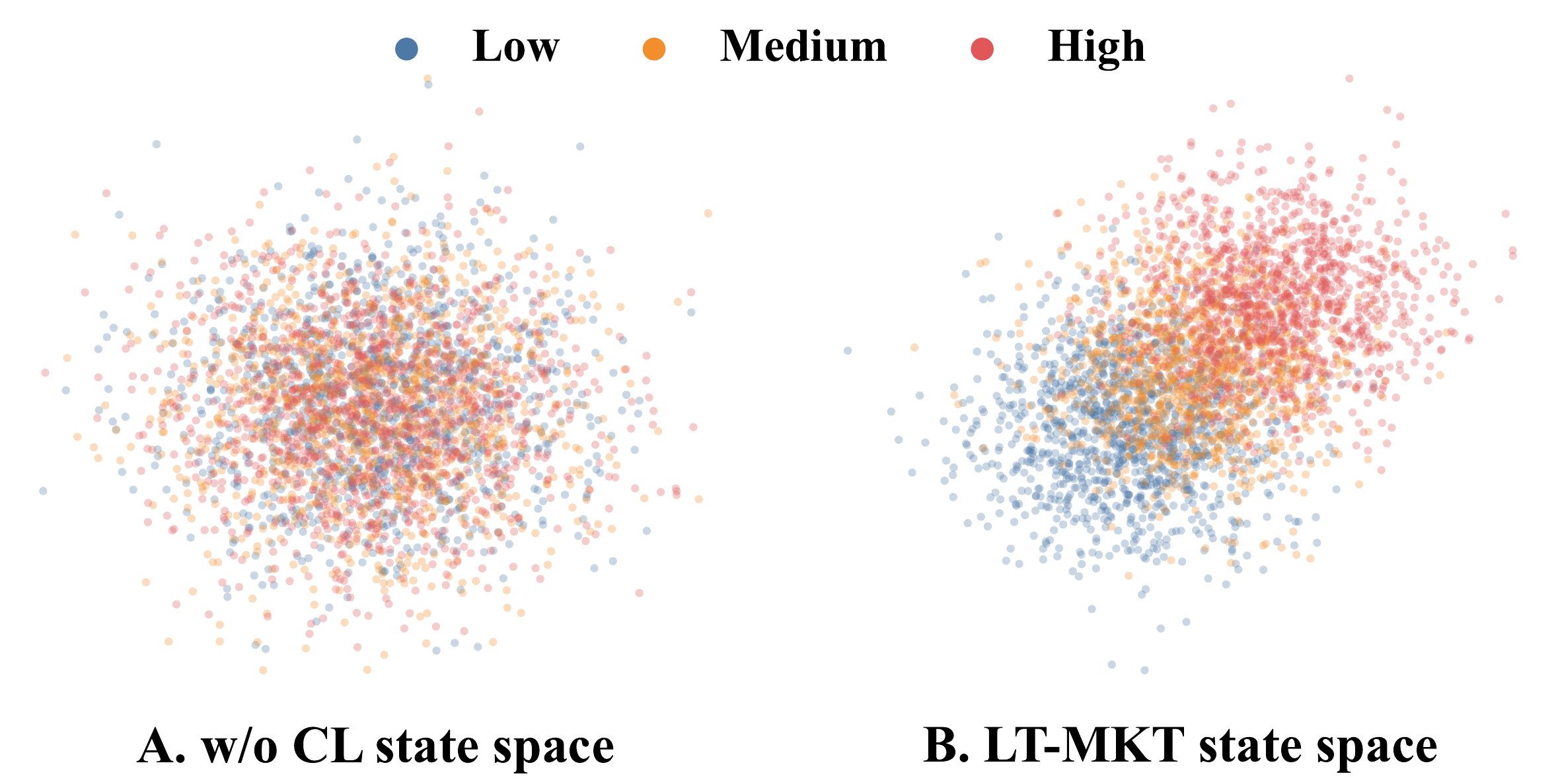}
    \vspace{-0.2cm}
  \caption{ Analysis of cognitive-load representations.}
  \label{fig:cl_probe}
  \vspace{-0.3cm}
\end{figure}

\subsection{Analysis of Cognitive Load Representation}

To further examine whether the cognitive load module learns meaningful load-aware representations, we conduct a representation-level analysis on the test interactions. For each interaction, we construct a cognitive load index (CLI) based on the three factors used in LT-MKT, namely question difficulty, domain transition, and domain coverage (Eq.~\ref{eq:clt}). Specifically, each factor is normalized using the training set statistics, and the final CLI is computed as the sum of the normalized values. We then divide test interactions into low-, medium-, and high-load groups based on CLI tertiles.
We compare the full LT-MKT model with its variant without cognitive load modeling, denoted as LT-MKT w/o CL. For both models, we extract the learned student state representations before the prediction layer and visualize them using UMAP~\cite{mcinnes2018umap}.

As shown in Figure~\ref{fig:cl_probe}, the representations learned by LT-MKT w/o CL are highly mixed across different cognitive-load groups, suggesting that the model does not explicitly organize student states according to the load structure of multi-domain learning. In contrast, LT-MKT produces a clearer and more continuous low-to-high load gradient in the representation space. This pattern indicates that the proposed cognitive load module helps encode cross-domain learning burden into the student state representation, rather than only acting as an additional input feature.

\begin{figure}[t]
  \centering
  \includegraphics[width=0.98\linewidth]{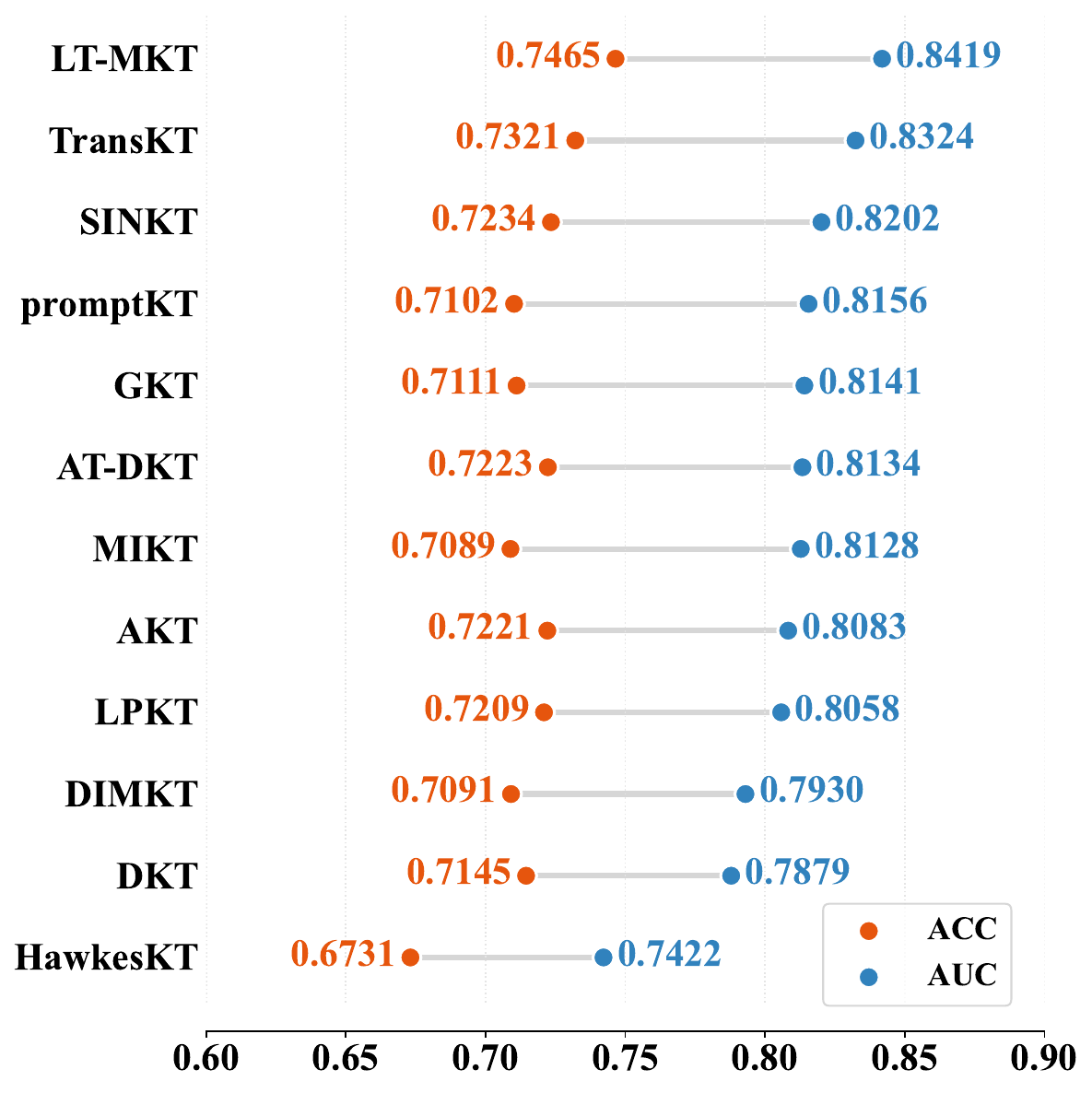}
    \vspace{-0.2cm}
  \caption{ Results in cold start scenarios.}
  \label{fig:data}
  \vspace{-0.4cm}
\end{figure}

\subsection{Performance under Cold-Start Scenarios}
Benefiting from the constructed Multi-domain Hierarchical Graph, LT-MKT is expected to evolve the knowledge states of KCs through their related concepts within and across domains. To more intuitively investigate the effectiveness of such knowledge transfer, we construct a cold-start scenario on the \textbf{JuniorH} dataset and evaluate the model's generalization ability on previously unseen concepts.
Specifically, the dataset is manually divided into training and testing sets, where the testing set contains 33 concepts that do not appear in the training set, accounting for 19\% of all concepts. Under this cold-start setting, the training and testing sets account for 84\% and 16\% of the interaction records, respectively.

Figure~\ref{fig:data} presents the performance of different methods in terms of AUC and ACC, from which several important observations can be made. First, LT-MKT consistently outperforms all other KT methods, demonstrating that modeling knowledge transfer within and across domains can effectively alleviate the cold-start problem, even when target concepts are absent from the training set. Second, methods considering knowledge structures (e.g., TransKT, SINKT, and GKT) also achieve relatively strong performance, suggesting that explicitly modeling relationships among concepts is beneficial for improving the generalization ability of KT models.

\begin{figure*}[t!]
  \centering
  \includegraphics[width=0.98\linewidth]{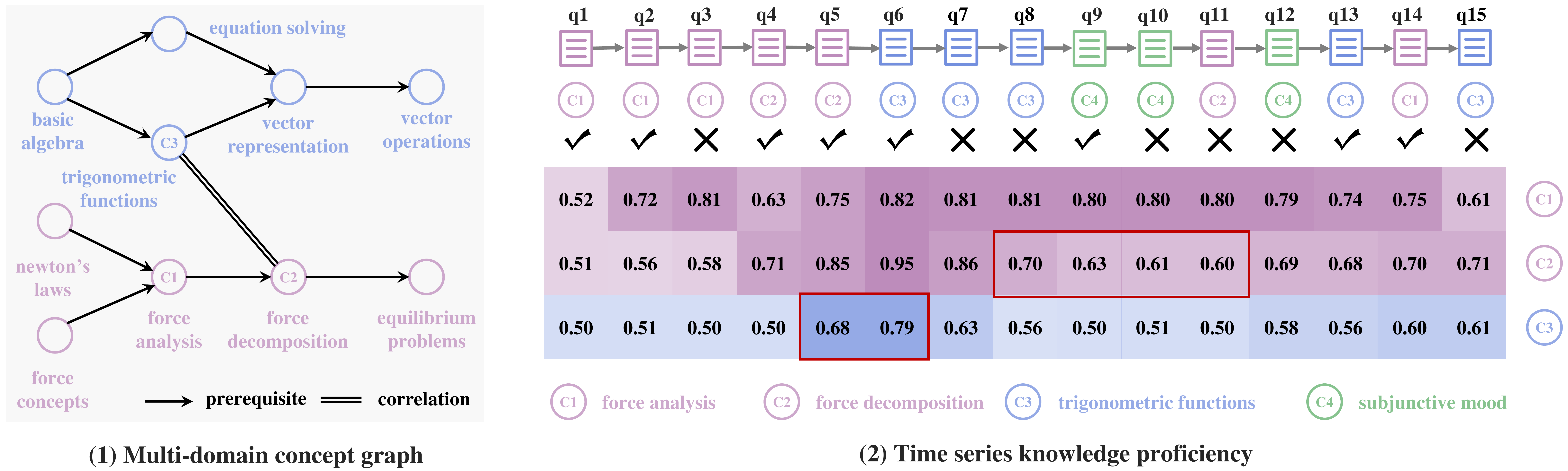}
  \vspace{-0.25cm}
  \caption{A case study of a student's cross-domain knowledge state evolution under LT-MKT.}
  \label{fig:Case Study}
  \vspace{-0.35cm}
\end{figure*}

\begin{figure}[t]
  \centering
  \includegraphics[width=0.95\linewidth]{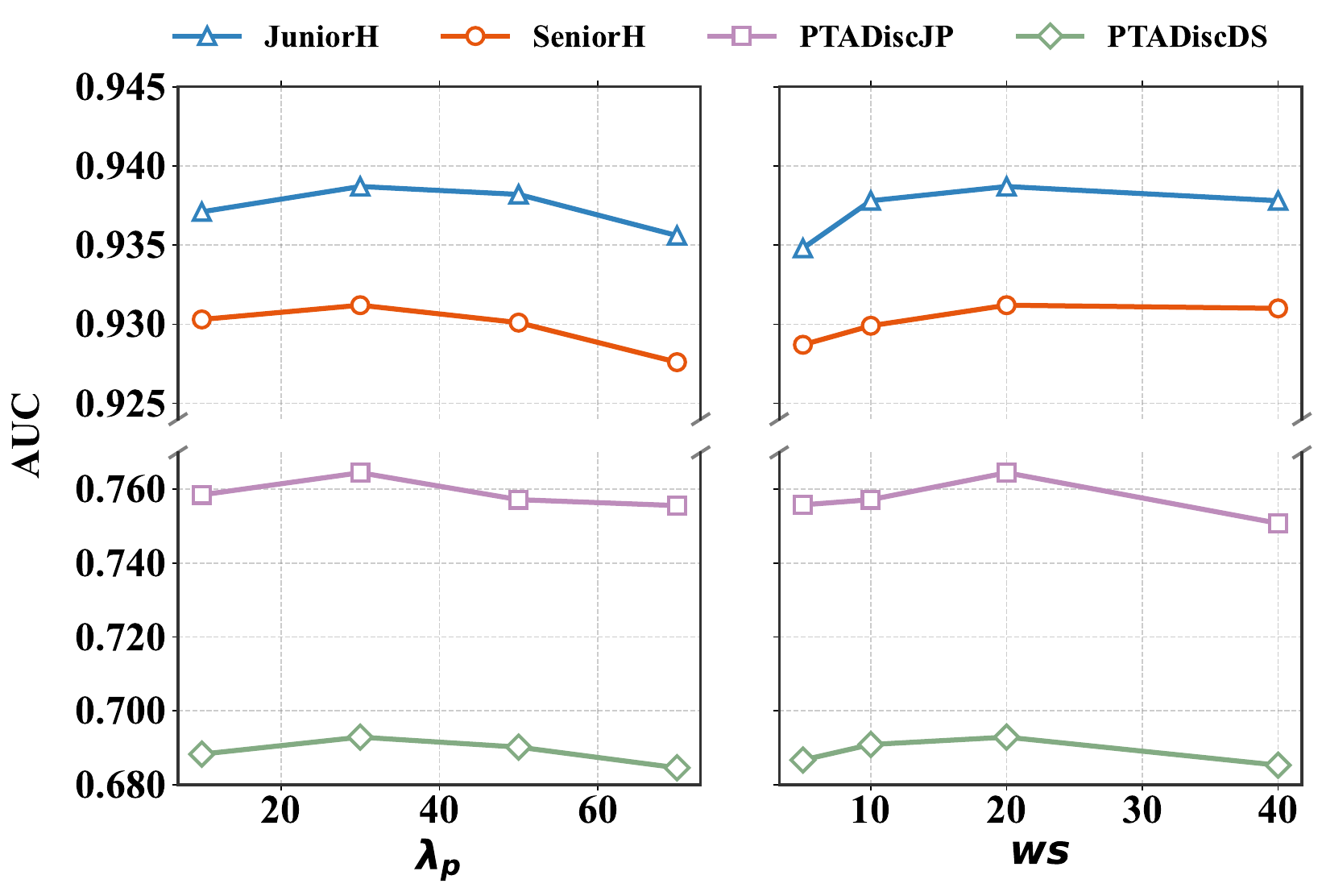}
  \vspace{-0.4cm}
  \caption{Parameter sensitivity analysis of $\lambda_P$ and $ws$.}
  \label{fig:Parameter Sensitivity Analysis}
  \vspace{-0.5cm}
\end{figure}

\subsection{Parameter Sensitivity Analysis}\label{sec:ws}

In this section, we analyze the sensitivity of two key hyperparameters, $\lambda_P$ in Eq.~\ref{eq:diff} and $ws$ in Eq.~\ref{eq:dt}.
We vary $\lambda_P$ within $\{10, 30, 50, 70\}$ while fixing $ws=20$, and vary $ws$ within $\{5, 10, 20, 40\}$ while fixing $\lambda_P=30$.
As shown in Figure~\ref{fig:Parameter Sensitivity Analysis}, LT-MKT achieves the best overall performance when $\lambda_P$ is set to a moderate value. When $\lambda_P$ is too small, questions with different empirical difficulty levels are compressed into coarse categories, making it difficult for the model to distinguish fine-grained cognitive load caused by question difficulty. Conversely, an overly large $\lambda_P$ introduces sparse and noisy difficulty levels, where small variations in correctness rates may be over-amplified. This weakens the stability of difficulty embeddings and leads to degraded performance. These results suggest that question difficulty should be modeled with sufficient but not excessive granularity.
The results for $ws$ show a similar pattern. A very small window only captures immediate domain switches and may miss short-term cross-domain learning patterns accumulated over several recent interactions. In contrast, an overly large window incorporates distant historical interactions, which may dilute the influence of recent domain switching and introduce irrelevant temporal noise. The best performance is generally obtained around $ws=20$, indicating that cognitive load caused by domain transitions is mainly reflected within a limited recent context. Overall, the sensitivity analysis confirms the necessity of properly balancing difficulty granularity and temporal context when modeling cognitive load in multi-domain learning.

\subsection{Case Study}
To complement the quantitative results, we present a qualitative case study to examine whether LT-MKT captures knowledge transfer and cognitive load in multi-domain learning. As shown in Figure~\ref{fig:Case Study}, the left part presents a partial concept graph, and the right part shows the evolution of a representative student's knowledge states from the \textbf{SeniorH} dataset over a sequence of interactions.
When the student answers $q6$, which involves concept $c3$, the model assigns an increased state to $c3$ even though the student has limited direct practice on it. This is because the student has previously shown strong mastery of $c2$, which is correlated with $c3$ in the concept graph. In contrast, the student fails to answer $q{11}$, which involves $c2$, after frequent domain switching. This indicates that cognitive load can weaken immediate learning gains even for previously well-trained concepts. Overall, this case study provides intuitive evidence that LT-MKT can jointly model beneficial knowledge transfer and load-induced learning friction.

\section{Conclusion}
In this paper, we proposed LT-MKT, a multi-domain knowledge tracing framework that jointly models cognitive load and knowledge transfer. By constructing a multi-domain hierarchical graph, LT-MKT captures load effects from cross-domain learning behaviors and propagates knowledge states through intra- and inter-domain relations. Experiments on real-world datasets show that LT-MKT consistently outperforms representative KT baselines, demonstrating its effectiveness in modeling students' knowledge states in multi-domain learning scenarios.
In future work, we will try to incorporate richer cognitive signals to further improve model performance, and investigate the application of LT-MKT in broader personalized learning and intelligent tutoring scenarios.

\begin{acks}
This work was supported by grants from the National Key Research and Development Program of China (Grant No. 2024YFC3308200), the National Natural Science Foundation of China (Nos. U25B2072 and 62477044), the Key Technologies R\&D Program of Anhui Province (No. 202423k09020039), the Young Elite Scientists Sponsorship Program by CAST (No. 2024QNRC001), and the Fundamental Research Funds for the Central Universities.
\end{acks}

\section*{GenAI Usage Disclosure}

During the preparation of this manuscript, generative AI tools such as ChatGPT and Grammarly were used strictly for grammar correction and sentence refinement. This paper does not contain any text generated entirely by large language models. All original ideas, experimental designs, and data analyses were conceived and conducted exclusively by the authors. Finally, the authors thoroughly reviewed all edited content and take full responsibility for the final version of the manuscript.

\bibliographystyle{ACM-Reference-Format}
\balance
\bibliography{sample-base}


\end{document}